%% file: iclr2026_conference.tex
\documentclass{article} %
\usepackage{iclr2026_conference,times}

\input{math_commands.tex}

\usepackage{graphicx}
\usepackage{hyperref}
\usepackage{url}

\definecolor[named]{ACMDarkBlue}{cmyk}{1,0.58,0,0.21}
\hypersetup{colorlinks,
  linkcolor=ACMDarkBlue,  %
  citecolor=ACMDarkBlue,  %
  urlcolor=ACMDarkBlue,   %
  filecolor=ACMDarkBlue}

\title{VisualScratchpad: Inference-time Visual Concepts Analysis in Vision Language Models}

\author{
Hyesu Lim$^{\dagger\ddagger}$\thanks{This work was done during internship at NAVER AI Lab. ${^{c}}$ Correspondence author.} ,
Jinho Choi$^{\dagger}$, 
Taekyung Kim$^{\ddagger}$, 
Byeongho Huh$^{\ddagger}$, \\
\textbf{ Jaegul Choo$^{\dagger}$, 
Dongyoon Han$^{\ddagger, c}$} \\
$^{\dagger}$KAIST AI, $^{\ddagger}$NAVER AI Lab
\vspace{-2em}
}

\iclrfinalcopy %
\begin{document}

\maketitle

\begin{abstract}
High-performing vision language models still produce incorrect answers, yet their failure modes are often difficult to explain.
To make model internals more accessible and enable systematic debugging, we introduce VisualScratchpad, an interactive interface for visual concept analysis during inference. We apply sparse autoencoders to the vision encoder and link the resulting visual concepts to text tokens via text-to-image attention, allowing us to examine which visual concepts are both captured by the vision encoder and utilized by the language model. VisualScratchpad also provides a token-latent heatmap view that suggests a sufficient set of latents for effective concept ablation in causal analysis. Through case studies, we reveal three underexplored failure modes: limited cross-modal alignment, misleading visual concepts, and unused hidden cues. Demos are available \href{https://iclrworkshop-visualscratchpad.github.io/visual_scratchpad_projectpage/}{here}.
\end{abstract}

\section{Introduction}

\begin{figure}[h]
    \centering
    \includegraphics[width=.9\linewidth]{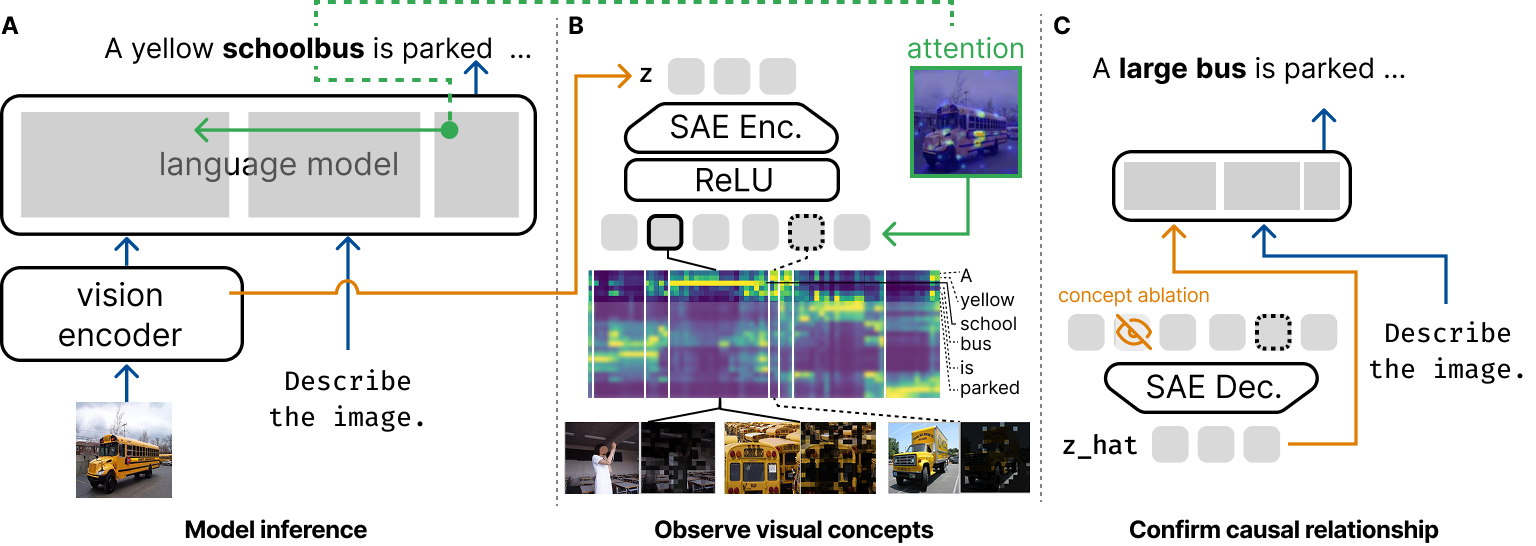}
    \vspace{-0.5em}
    \caption{\textbf{VisualScratchpad pipeline.} \textbf{A.} During inference in a vision-language model, we extract the intermediate representation \texttt{z} from the vision encoder. \textbf{B.} A sparse autoencoder processes \texttt{z} to produce concept activations. The attention map from output text tokens to image tokens is applied at the patch level to weight these activations. Latents exhibiting similar activation patterns across output tokens are then clustered and visualized in a token–latent heatmap. \textbf{C.} The causal influence of these concepts on the model’s output can be evaluated through latent ablation.}
    \label{fig:overview}
    \vspace{-0.5em}
\end{figure}

Why do vision-language models (VLMs) make errors? Do the failures come from perceiving insufficient visual cues or relying on misleading ones? Although this question appears simple, answering it rigorously is challenging due to the limited interpretability of current VLMs. Mechanistic interpretability aims to address such questions through analyzing model weights and their activation patterns, and by reverse-engineering the functional roles of individual neurons~\citep{bereska2024mechanistic, sharkey2025open}. 

A central challenge is that a single neuron is often activated by multiple unrelated concepts~\citep{elhage2022toy}. Sparse autoencoders (SAEs)~\citep{ng2011cs294a, ranzato2007sparse} mitigate this issue by expanding dense representations into a higher-dimensional sparse basis. In the last couple of years, SAEs have shown their potential in large language models~\citep{cunningham2023sparse}, vision encoders~\citep{limsparse}, diffusion models~\citep{surkov2025unpacking}, and more domain-specific foundation models, including those for medical imaging~\citep{dasdelen2025cytosae}. SAEs decompose representations into granular units that can be assigned semantic meaning through reverse engineering, typically by examining the inputs that strongly activate them. These sparse units are known to be more interpretable than the original model representation. 
Although recent work has achieved technical improvements in SAE architectures and training strategies~\citep{bussmann2024batchtopk, fel2025archetypal}, practical applications that effectively leverage SAE latents remain at an early stage, in part because the community still lacks an interface that supports a systematic analysis pipeline. Nevertheless, a growing body of research has begun to illustrate the promise of SAEs, for AI safety~\citep{bereska2024mechanistic}, scientific discovery~\citep{wang2026alzheimers}, explaining mechanisms for adaptation to downstream tasks~\citep{limsparse}, and enabling scalable auto-annotation of dataset analysis~\citep{choi2025conceptscope}. Motivated by these emerging directions, we introduce an interface that provides a unified and practical pipeline for SAE-based concept exploration, inference-time debugging, and causal analysis.

In this work, we apply SAEs to understand and debug inference-time behavior of vision-language models (VLMs). Specifically, we aim to discover which visual concepts are captured by the vision encoder and subsequently used by the language model. To this end, we introduce \textbf{VisualScratchpad}, an interactive interface that supports inference-time concept inspection and causal testing through latent ablation. To avoid confounds introduced by projection layers and cross-modal attention when interpreting visual tokens inside the language model, we propose to apply SAEs directly to the vision encoders and link visual concepts to language tokens via text-to-image attention maps. We verify causal influence by replacing a latent's activation with a user-specified value. However, because SAE latents vary in semantic granularity~\citep{bricken2023monosemanticity}, selecting an appropriate subset for effective concept steering can be challenging. To make this selection process more tractable, we introduce a token–latent activation heatmap, where the latents are clustered and visualized based on their activation similarity across output tokens.

Lastly, we share several findings that address our research question of why VLMs produce incorrect responses through three case studies. The results highlight the potential and use cases of our proposed VisualScratchpad interface. First, we observe that even when the correct cue was captured by the vision encoder, the model may fail to utilize it; modifying the input prompt with a more direct description corrected the output. Second, when the model relies on a misleading cue, removing it changes the prediction. Third, although the vision encoder may capture multiple plausible visual concepts, the model often relies only on the most dominant one. Adjusting the relative activation strengths flips the output. These findings highlight key failure modes in VLMs and demonstrate how VisualScratchpad can support systematic debugging.

\vspace{-1em}
\section{Method}
\vspace{-.5em}

\noindent \textbf{Extracting visual concepts.}
Following \cite{limsparse}, we train vanilla SAEs on frozen CLIP-ViT-large~\citep{radford2021learning}
with an expansion factor of 32. Image token representations $z\in\R^{d_{\text{model}}}$ are used as the input of the SAE, and the 1024-dimensional hidden representations are expanded into 32,768 latents $h\in\R^{d_{\text{SAE}}}$, i.e., $h=\text{ReLU}(W_{\text{enc}}^\top z), \hat{z}= W_\text{dec}^\top h.$ After training, both train and validation images are passed through the SAE to collect maximally activating images, from which activation frequency and mean activation strength are computed. More details are provided in \S~\ref{app:sae_training_details}.

\noindent \textbf{Linking visual concepts via text-to-image attention.}
We analyze VLMs following the typical architecture, where the vision encoder produces image tokens that are concatenated with text tokens, and text tokens attend to image tokens through cross-attention. This architecture has been utilized by multiple popular model families, including LLaVA~\citep{liu2023visual}, LLaMA~\citep{grattafiori2024llama}, Qwen~\citep{bai2025qwen2}, InternVL~\citep{chen2024internvl}, and Gemma~\citep{team2024gemma}. We refer to the attention weights from a single text token to all image tokens (patches) as an attention map. We average these weights over all layers and attention heads, resulting in $\rm{attn}\in \R^{n_{\text{patches}}}$. This attention map indicates which image regions the model focuses on when processing that specific text token, with a higher value for stronger attention.

\begin{figure}[t]
    \centering
    \includegraphics[width=.9\linewidth]{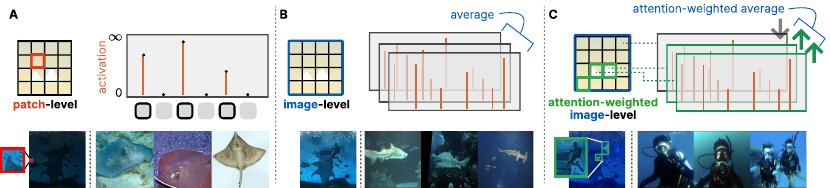}
    \vspace{-1em}
    \caption{\textbf{Attention-based concept re-ranking}. \textbf{A.} SAEs return latent activations for each image patch. \textbf{B.} Image-level activations can be computed by naïvely averaging activations across all patches, or \textbf{C.} by applying a weighted average where the text-to-image attention map serves as the weighting coefficient, promoting concepts relevant to the text tokens to the top of the ranking. The bottom row shows the top-ranked concept obtained from the corresponding method.}
    \label{fig:method}
    \vspace{-1.5em}
\end{figure}

To link visual concepts in the vision encoder with the language model, we multiply the attention map with SAE latent activations $\mathbf{h}\in\R^{n_{\text{patches}} \times d_{\text{SAE}}}$ in a patch-wise manner, i.e., $\mathbf{h}^\top\rm{attn}$. Attention-weighted averaging re-ranks the activated visual concepts by pushing latents from attended regions toward the top of the ranking, while down-ranking concepts originating from less attended regions. Figure~\ref{fig:method} illustrates the \textbf{attention-based concept re-ranking} and example results.

\noindent \textbf{Testing causal influence.} 
To verify whether the identified visual concepts exert causal influence on the model’s generated outputs, we compare the original output with the output obtained after zeroing out the activations of selected latents. Because SAE latents vary in semantic granularity, they often exhibit hierarchical or correlated relationship~\citep{chanin2024absorption, wittenmayer2026insight}. As a result, a concept may reappear through another interconnected latent even after one latent is ablated. Identifying a group of latents corresponding to a single concept is therefore crucial for performing effective concept ablation.

To support this, we construct a \textbf{token–latent activation heatmap}. We first remove noisy latents based on activation statistics (\S~\ref{app:heatmap_vis}; Figure~\ref{chapter4:fig:sae_stats}), then select, for each text token, the top-$k$ ($k=20$) latents ranked by attention-weighted activation strength. This filters out non-activated latents and yields a more interpretable, user-friendly view. Taking the union across tokens produces a set of meaningfully activated latents that are likely to carry causal influence. Depending on the number of tokens, the resulting heatmap ranges from size $n \times (n\cdot k)$ to $n \times k$ for $n$ tokens. We cluster these latents based on their activation similarity across tokens, and normalize each column for clearer visualization (\S~\ref{app:heatmap_vis}; Figure~\ref{fig:heatmap_method}).

In Figure~\ref{fig:causal_scuaba_ex}, for instance, we identify four major semantic topics within the generated caption and their corresponding latent clusters. Ablating each topic removes only the associated portion of the output while leaving unrelated content intact, demonstrating that clusters based on token-wise activation similarity indeed show causal influence on the model’s predictions.

\noindent \textbf{Interface.} We introduce VisualScratchpad, an interactive interface for inference-time visual concept inspection and ablation. 
As illustrated in Figure~\ref{chapter4:fig:overview}, The interface consists of four main components: SAE latent exploration, model inference, internal observation, and internal modification.
The exploration panel provides statistics of latent activations in the training or validation dataset and the decoder weight based U-map view of latent clusters, which allows overall understanding of learned concepts.
Model inference of user-provided input supports the vision-question answering style or CLIP-style classification tasks. 
Internal observation shows attention maps, token-wise latent activation, token-latent heatmap, and selected latent's input mask with reference images.
By analyzing this, users can confirm their causal relationship through concept steering.
Demos are available \href{https://iclrworkshop-visualscratchpad.github.io/visual_scratchpad_projectpage/}{here}.

\vspace{-.5em}
\section{Case studies: Revealing three failure modes}
\vspace{-.5em}

\begin{figure}[t]
    \centering
    \includegraphics[width=1\linewidth]{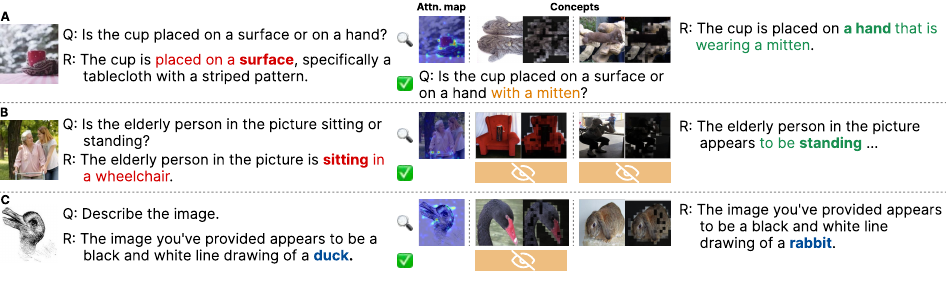}
    \vspace{-1.5em}
    \caption{\textbf{Case studies.} \textbf{A.} Limited cross-alignment, \textbf{B.} misleading visual cues, and \textbf{C.} unused hidden cues can cause wrong answer.}
    \label{fig:case_studies}
    \vspace{-1.5em}
\end{figure}

We present three failure cases where LLaVA-Next-8B~\citep{liu2024llavanext} initially produces incorrect outputs, but analysis with VisualScratchpad allows us to debug and correct its behavior. For Case 1 and Case 2, we use examples from the MMVP dataset~\citep{tong2024eyes}. We identify two distinct failure modes: in Case 1, the model detects the correct visual cues but fails to utilize them, whereas in Case 2, the model is misled by incorrectly detected cues. Although the dataset is designed to expose the perceptual limitations in current vision encoders for VLMs~\citep{tong2024eyes}, our results illustrate how model failures can stem from either underusing relevant information or over-relying on spurious signals. Motivated by these observations, we pose a further question: What happens when there are multiple plausible visual cues that could all be correct? To explore this, we design Case 3 using an optical illusion, where competing interpretations coexist. Details are in \S~\ref{app:case_studies}.

\noindent \textbf{Case 1: Limited cross-modal alignment.}
Figure~\ref{fig:case_studies} \textbf{A} shows an example where the model is asked whether a cup is placed on a surface or a hand. The original response was incorrect, stating that the cup is on the surface. The model incorrectly answers “on a surface,” even though the corresponding attention map shows it attends to the hand and activates concepts like (knitted) gloves. We hypothesize that the textual concept ``hand'' is not directly aligned with the visual concept of ``gloves,'' leading the model to underutilize the correct cue. After rephrasing the question to include more detail (``Is the cup placed on a surface or on a hand \textit{with a mitten}?''), the model produces the correct answer. This suggests that the relevant visual concept exists in the representation but can be poorly aligned with its linguistic counterpart.

\noindent \textbf{Case 2: Grounding on misleading cues.}
In Figure~\ref{fig:case_studies} \textbf{B}, the question asks whether an elderly person is sitting or standing. The model incorrectly predicts ``sitting''. Although the model attends to the ``walker'' in the image, concepts related to a ``(wheel)chair'' and ``sitting'' are activated. Removing these concepts flips the response to ``standing.'' This case illustrates a common failure mode in VLMs: reliance on associative yet semantically inappropriate cues.

\noindent \textbf{Case 3: Unused hidden cues.}
Figure~\ref{fig:case_studies} \textbf{C} examines a case with multiple plausible interpretations. Given an optical-illusion image, the model initially describes a ``duck', though rabbit-related concepts are also activating. Ablating duck-related latents and amplifying rabbit-related ones shifts the output to describing a ``rabbit''. This demonstrates that VLMs may internally encode richer visual information than what appears in their final output, shedding light on their inconsistent behaviors.

\vspace{-1em}
\section{Related work}
\vspace{-.5em}
\noindent \textbf{SAEs for VLMs.} 
SAE-V~\citep{lou2025sae} applies an SAE to an intermediate language-model layer of VLMs, where cross-attention integrates text and vision features, and uses the resulting latents to assess dataset quality in terms of cross-modal alignment. VL-SAE~\citep{shen2025vl} proposes a method to align and interpret concepts from the vision and text encoders. In contrast, VisualScratchpad links visual concepts from the vision encoder to text tokens \textit{post hoc} via attention maps, enabling analysis across different cross-modal linking mechanisms while retaining direct control over the vision encoder.

\noindent \textbf{Interface for SAEs.}
Only a few interactive tools exist for working with SAEs. Neuronpedia~\citep{neuronpedia} and ConceptViz~\citep{li2025conceptviz} provide interfaces for concept analysis and latent steering but are limited to language models. PatchSAE~\citep{limsparse} offers a demo for inspecting vision-model concepts but does not support latent steering or VLM inference. VisualScratchpad provides a unified interface for multimodal analysis, integrating concept inspection, steering, and inference-time debugging for VLMs.

\vspace{-1em}
\section{Conclusion} \label{chapter3:sec:discussion}
\vspace{-.5em}
In summary, we examine pure visual concepts by applying SAEs directly to the vision encoder and inspect output-specific concepts through attention-based linking. To support effective concept steering, we introduce a token–latent heatmap view that helps users identify sufficient and relevant latents. Together, these components form an interactive interface that unifies inference-time analysis and general SAE-based concept exploration. Through case studies, we show that VLM failures can stem from limited cross-modal alignment, misleading concepts, or unused hidden cues.

\subsubsection*{Limitation and outlook} Connecting vision SAEs to language models does not explicitly capture how image-token representations are transformed inside the language model during cross-modal processing. In addition, the interface is built for interactive analysis and does not directly scale to large, fully automated experiments. Nonetheless, the approach provides a practical framework for dissecting VLM behaviors and opens the door to future extensions, such as integrating deeper cross-modal tracing, automating large-scale causal analyses, and applying similar techniques to broader multimodal architectures.

\subsubsection*{Author Contributions}
    \textit{Conceptualization:} HsL with comments from TkK, BhH, JgC, and DyH; 
    \textit{Methodology:} HsL, DyH;
    \textit{Software:} HsL, JhC;
    \textit{Formal analysis:} HsL, JhC;
    \textit{Investigation:} HsL;
    \textit{Writing--Original Draft:} HsL.

\subsubsection*{Acknowledgments}
This work is supported by NAVER AI Lab and NAVER Smart Machine Learning (NSML) platform \citep{sung2017nsml}.

\bibliography{iclr2026_conference}
\bibliographystyle{iclr2026_conference}

\newpage
\appendix
\section{Appendix}

\subsection{SAE training details}
\label{app:sae_training_details}

We follow the PatchSAE training setup introduced in \citet{limsparse}, using CLIP-ViT-large\footnote{\url{https://huggingface.co/openai/clip-vit-large-patch14}} as a vision backbone and a vanilla sparse autoencoder (SAE) trained with mean-squared error reconstruction loss and an L1 sparsity penalty of $8\times10^{-5}$. The SAEs are trained on ImageNet-1K for 2M steps with a learning rate of $1\times10^{-4}$ with decay. CLIP-ViT-large has 24 layers in total  (layer.0 from layer.23), and we extract activations from layers 2, 6, 10, 14, 18, and 22 (the penultimate layer used as the image feature in LLaVA-Next-8B~\citep{liu2024llavanext}). 

\begin{figure}[h]
    \centering
    \includegraphics[width=1.\linewidth]{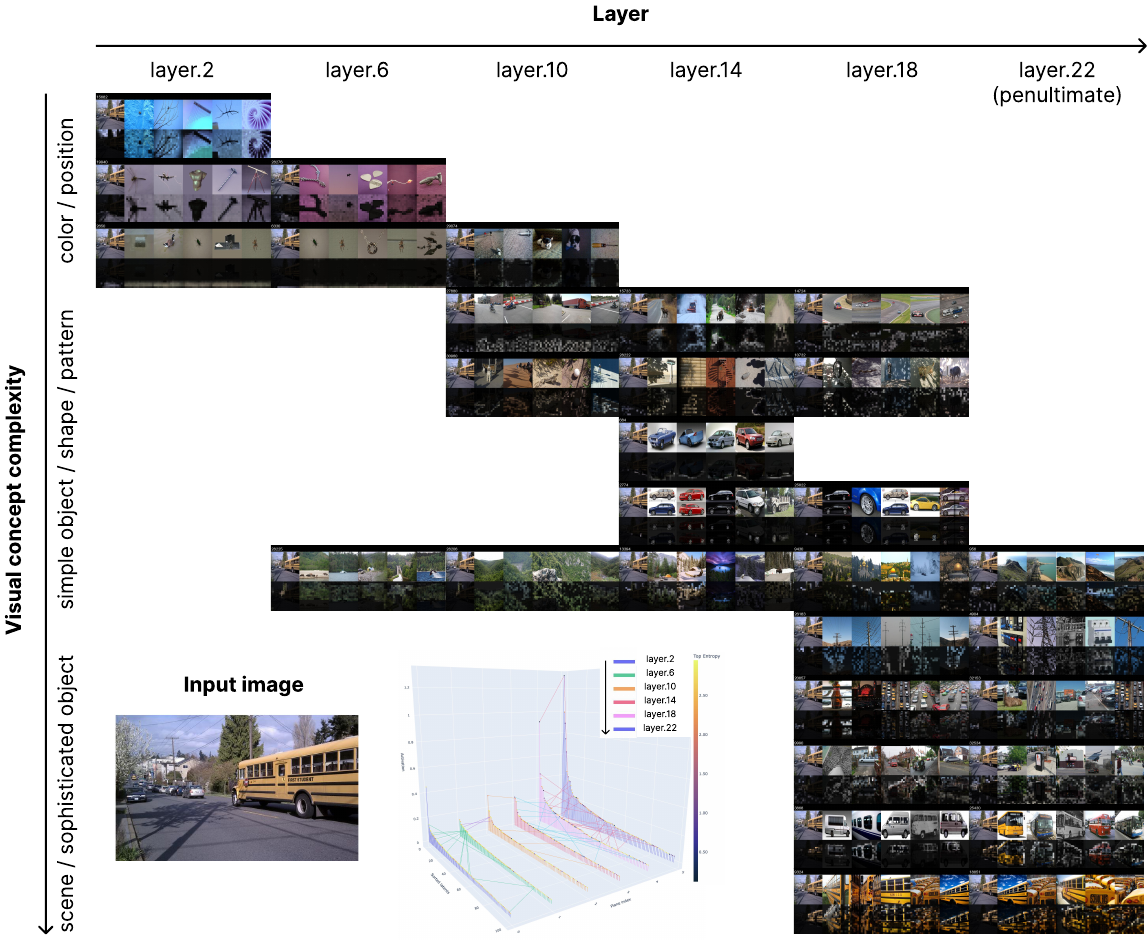}
    \caption{\textbf{Complexity of visual concepts across layers.} Early layers capture simple visual components, including position and color, mid layers capture object-level shape and patterns, and late layers understand scene-level events.}
    \label{fig:layer_comparison}
\end{figure}

In accordance with the observation from previous work~\citep{limsparse}, across layers, we observe a progression in the learned concepts: early layers capture colors and simple textures, middle layers encode object-level representations, and deeper layers focus on scene-level semantics. For our case studies, we use layer.18 due to its strong alignment with object-centric concepts. Figure~\ref{fig:layer_comparison} shows examples of activated latents across layers. Trained checkpoints are released in our \href{https://iclrworkshop-visualscratchpad.github.io/visual_scratchpad_projectpage/}{codebase}.

\subsection{Heatmap visualization and filtering details}
\label{app:heatmap_vis}
\begin{figure}[h]
    \centering
    \includegraphics[width=1.\linewidth]{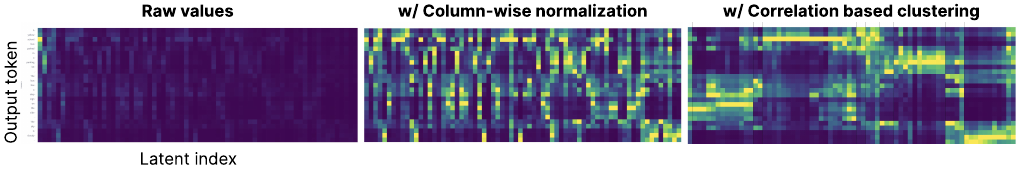}
    \caption{\textbf{Token-Latent heatmap visualization.} Raw values are difficult to analyze, so we normalize in column-wise to show if the latent is specifically attended by certain tokens or by overall tokens. Moreover, we cluster and sort by activation correlation in column-wise, using hierarchical clustering.
    }
    \label{fig:heatmap_method}
\end{figure}

\begin{figure}[h]
    \centering
    \includegraphics[width=1.\linewidth]{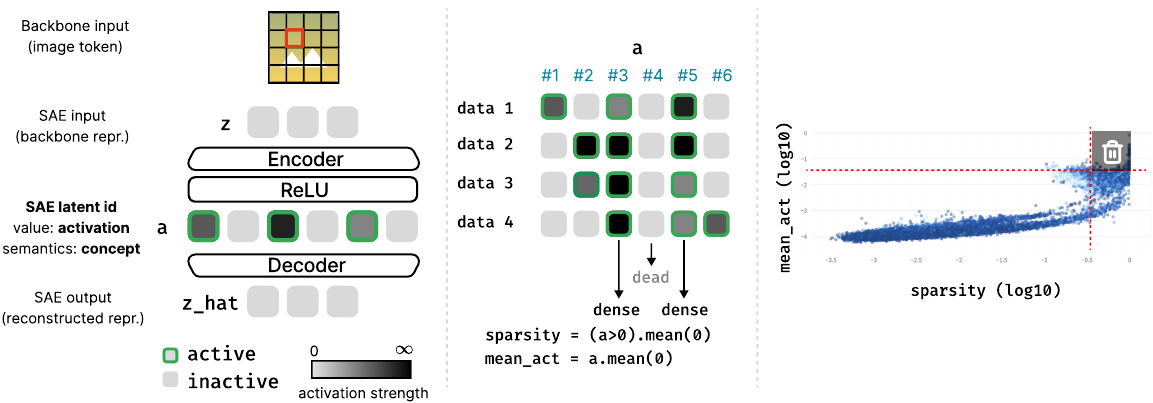}
    \caption{\textbf{Illustration on how SAE latent statistics are computed.} After training the SAE, we pass the dataset through the model and compute each latent's mean activation and sparsity, yielding a compact summary of its activation distribution. \textbf{Right:} Based on the mean activation and sparsity, we filter out latents in top $k$\% of each axis treating them as noisy latents.
    }
    \label{chapter4:fig:sae_stats}
\end{figure}

For the heatmap visualization, we apply column-wise normalization and correlation-based clustering (Figure~\ref{fig:heatmap_method}). The interface offers multiple configuration options, including no normalization, row-wise normalization, or column-wise normalization, combined with either hierarchical or k-means clustering using correlation or Euclidean distance. In addition, we leverage latent statistics to filter out noisy latents, removing the top 2\% based on mean activation and activation frequency. Figure~\ref{chapter4:fig:sae_stats} illustrates the computation of these statistics.

\subsection{Causal Analysis}

\begin{figure}[h]
    \centering
    \includegraphics[width=1.\linewidth]{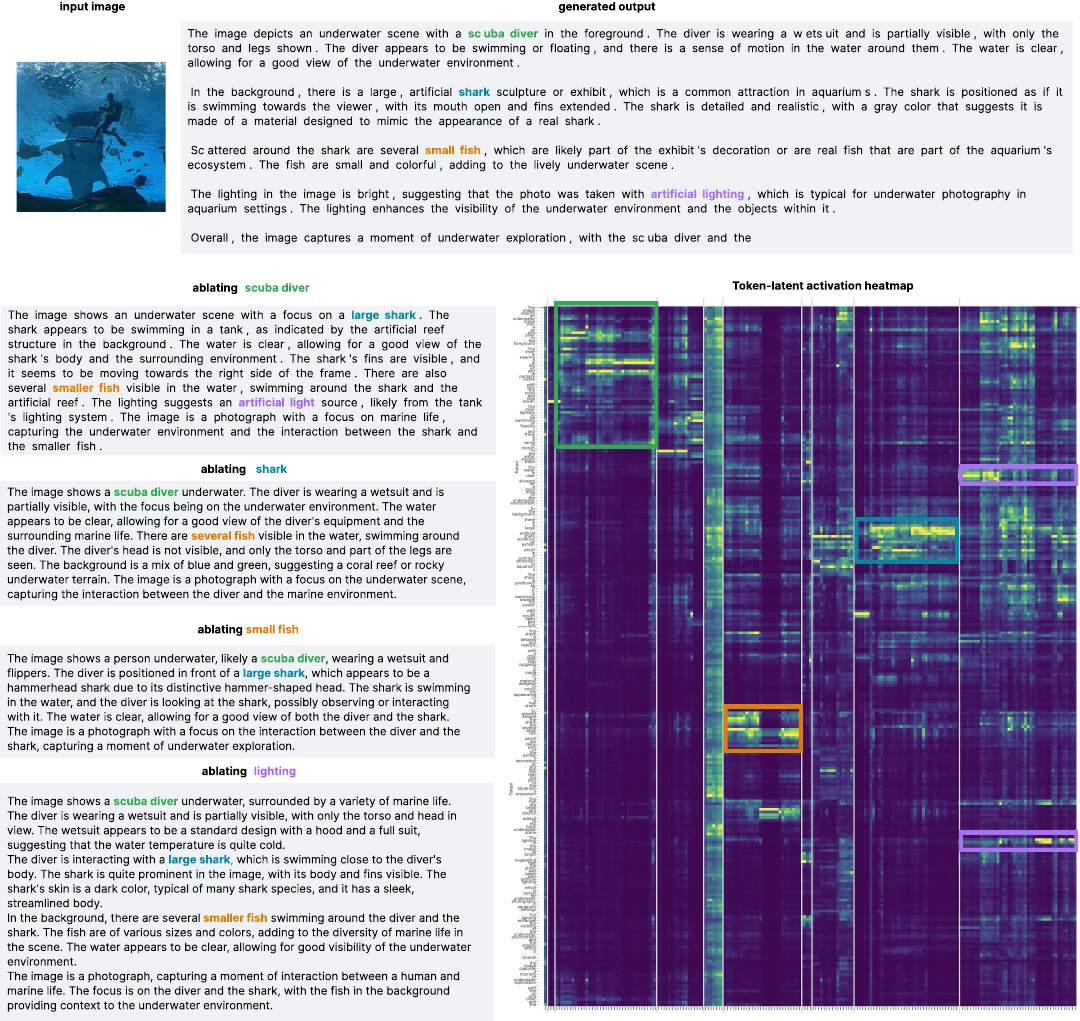}
    \caption{\textbf{Causal analysis example.} Given an image and a prompt to describe it, the VLM generates a caption and VisualScratchpad produces a token–latent heatmap. Latents are re-ranked using token attention and clustered based on activation similarity. Each cluster roughly corresponds to a main topic in the caption. By zeroing out the latents in each cluster and reconstructing the representations, we observe how the outputs change relative to the original. Ablating latent clusters that correspond to distinct semantic topics removes those topics from the generated caption, demonstrating their causal role in shaping the model’s predictions.}
    \label{fig:causal_scuaba_ex}
\end{figure}

To illustrate the effectiveness of the token–latent heatmap view for concept steering in causal analysis, we present an example in Figure~\ref{fig:causal_scuaba_ex}. Given an underwater image, the VLM’s caption contains four major semantic topics. The token–latent heatmap reveals four corresponding latent clusters. When each cluster is ablated, only the associated part of the caption disappears while unrelated content remains unchanged, demonstrating that these latent clusters exert causal influence on the model’s output.

\newpage
\subsection{Interface}
\label{app:interface}

\begin{figure}[t]
    \centering
    \includegraphics[width=1.\linewidth]{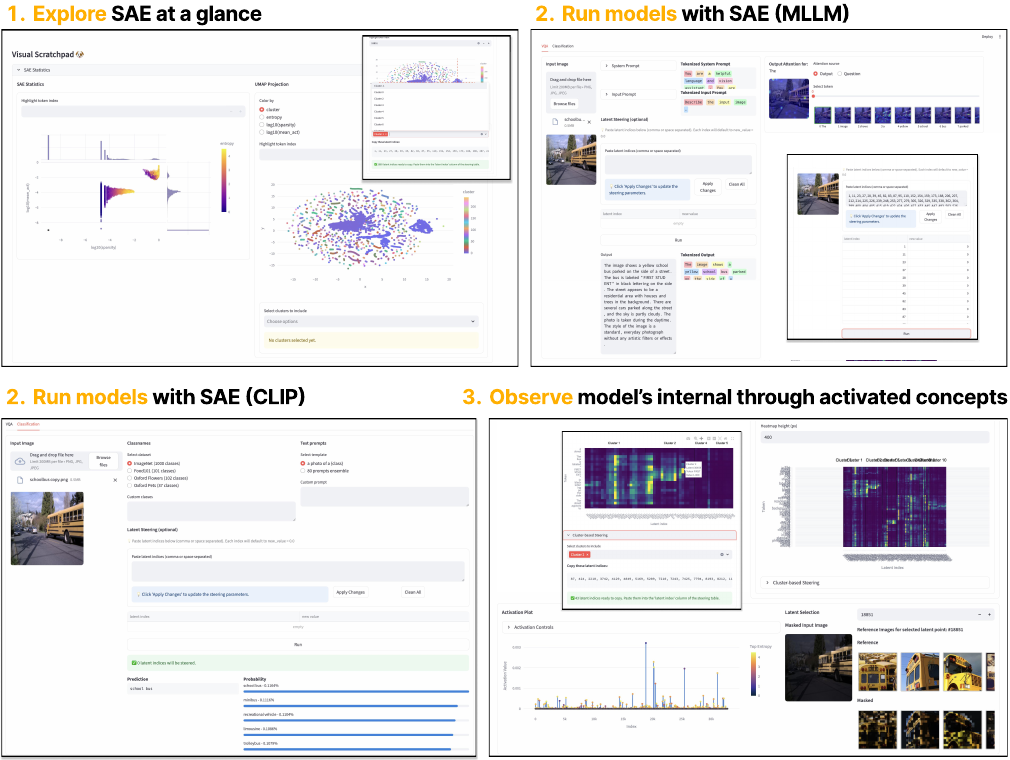}
    \caption{\textbf{Three main components of \texttt{VisualScratchpad}} include (1) exploring SAE latents, (2) model inference with SAE (vision-question answering using VLMs and image classification using CLIP), and (3) close-up observation on activated concepts.}
    \label{chapter4:fig:overview}
\end{figure}

\begin{figure}[t]
    \centering
    \includegraphics[width=1.\linewidth]{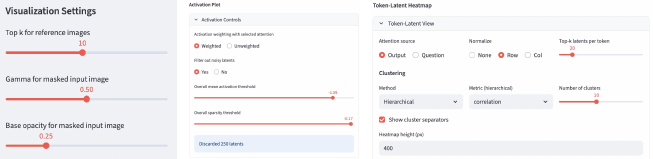}
    \caption{\textbf{Control pannels of VisualScratchpad.} Users can adjust visualization settings for spatial attribution masks, configure options for activation-value plots, and customize heatmap displays.}
    \label{chapter4:fig:controls}
\end{figure}

Figure~\ref{chapter4:fig:overview} shows the overview of our interface. Demo videos and codes are available in our project page\footnote{\url{https://iclrworkshop-visualscratchpad.github.io/visual_scratchpad_projectpage/}}.

\paragraph{Exploring SAE latents.} As shown in Figure~\ref{chapter4:fig:overview} (1), the top panel provides an overview of SAE latent statistics and concept clusters.
Mean activation and sparsity are computed from validation data, while label entropy is derived from the top-$k$ maximally activating samples (Figure~\ref{chapter4:fig:sae_stats}).
Decoder weights, which are considered as latent semantic directions, are visualized and clustered in a UMAP projection.
The statistics scatter plot offers a quick overview of how latents are distributed in terms of activation frequency, while the clusters reveal semantic structure and relationships between concepts.

\paragraph{Model inference: VQA using VLM.}
During VQA inference, the input image (and text prompt) is forwarded through both the backbone model and the SAE.
We additionally display the cross-attention map from each question or generated token to the image tokens, providing an intuitive view of where the model attends during generation (Figure~\ref{chapter4:fig:overview} (2)).
Because PatchSAE operates on patch-level embeddings aligned with the attention map, attention-weighted pooling highlights the most relevant concepts for each token.
To make this interpretable at scale, we display token-latent heatmap.
This allows users to follow the model's reasoning step-by-step and observe clusters of jointly activated concepts.

\paragraph{Model inference: Classification using CLIP.}
We also support CLIP-based image classification using the text encoder. (Figure~\ref{chapter4:fig:overview} (3))
Users can select from built-in example class lists or provide custom class names.
Prompt templates are editable, and classification outputs along with probabilities are displayed.

\paragraph{Model inference with steering.} 
VisualScratchpad enables steering by modifying the latent activations before decoding (Figure~\ref{chapter4:fig:overview} (4)).
By default, selected latents are ablated by setting their activations to zero, but users can alternatively emphasize targeted concepts by increasing activation values. 
UMAP-based latent clusters and token–latent heatmap clusters provide natural starting points for selecting latents to steer, allowing users to adjust specific concepts and observe the resulting changes in model outputs.

\paragraph{Observe active latents.}
During model forward passes, SAE activations are displayed as an activation bar (Figure~\ref{chapter4:fig:overview} (4)).
Selecting a latent reveals reference images together with their spatial attribution masks.
In VQA settings, the activation bar can be displayed either with attention-weighted activations or in an unweighted form.

\paragraph{Option control panels.} 
To facilitate focused exploration, VisualScratchpad provides filtering based on latent statistics, allowing users to exclude uninformative or low-impact latents. 
Latents that activate in most images are often difficult to interpret, since their maximally activating samples may contain multiple unrelated concepts. 
We therefore filter out latents with high activation frequency and high mean activation, which are more likely to reflect generic or mixed features. 
Each panel supports interactive hyperparameters, such as clustering settings, masking options, filtering modes, and weighting strategies, enabling fine-grained control over the visualization and analysis workflow (Figure~\ref{chapter4:fig:controls}).

\subsection{Case study}
\label{app:case_studies}

\paragraph{Dataset.} We use examples from the MMVP dataset~\citep{tong2024eyes}, which contains image pairs with high CLIP~\citep{radford2021learning} cosine similarity but low DINO~\citep{caron2021emerging} similarity, referred to as “CLIP-blind pairs.” These pairs capture cases where CLIP fails to distinguish subtle visual differences. The dataset includes questions focused on those fine-grained distinctions. We use MMVP images for Case 1 and Case 2 to analyze why VLMs fail to identify such differences, and for Case 3, we use a standard optical illusion image.

\paragraph{Experimental setup.} We perform concept ablation using latents from layer 18. When training SAEs on multiple layers of CLIP and comparing the resulting concepts, we observe that earlier layers primarily encode colors and simple patterns, mid-level layers capture object-level concepts, and later layers reflect scene-level understanding (\S~\ref{app:sae_training_details}; Figure~\ref{fig:layer_comparison}). 

\begin{figure}[h]
    \centering
    \includegraphics[width=1.\linewidth]{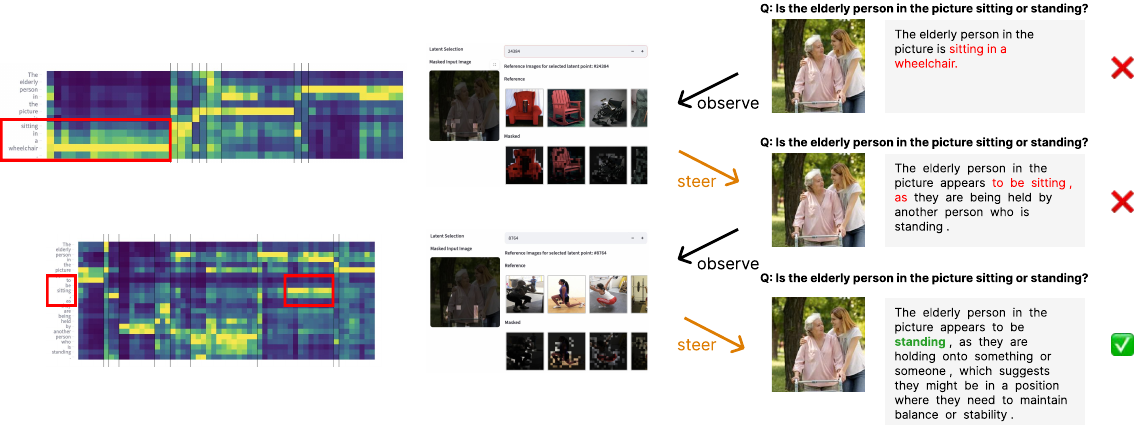}
    \caption{\textbf{Details for Case 2.} We iterated for two turns to remove sitting-related concepts.
    }
    \label{chapter4:fig:wheelchair}
\end{figure}

\paragraph{Details for Case 2.} In Case 2, we apply concept steering for two iterative rounds. Increasing the value of 
$k$ in the token–latent heatmap construction can reduce the number of required iterations. The full process is illustrated in Figure~\ref{chapter4:fig:wheelchair}.

\end{document}

%% file: math_commands.tex
\usepackage{amsmath,amsfonts,bm}

\def\eqref#1{equation~\ref{#1}}

\def\1{\bm{1}}

\DeclareMathAlphabet{\mathsfit}{\encodingdefault}{\sfdefault}{m}{sl}
\SetMathAlphabet{\mathsfit}{bold}{\encodingdefault}{\sfdefault}{bx}{n}

\newcommand{\R}{\mathbb{R}}